\documentclass[letterpaper, 10 pt, conference]{ieeetran}
\usepackage[letterpaper, left=0.75in, right=0.75in, bottom=0.75in, top=0.75in]{geometry}
\usepackage{graphicx}
\usepackage{times}
\usepackage{amsmath}
\usepackage{amssymb}
\usepackage{amsfonts}
\usepackage{array}
\usepackage{algorithm}
\usepackage{algorithmicx}
\usepackage[noend]{algpseudocode}
\usepackage{multirow,color}
\usepackage{algpseudocode}
\usepackage{varwidth}
\usepackage{subfig}
\usepackage{booktabs}
\usepackage{gensymb}
\usepackage[hyphens]{url}
\usepackage[export]{adjustbox}
\usepackage[font=footnotesize]{caption}
    
\usepackage[inline]{enumitem}
\usepackage{dblfloatfix}
\usepackage{xcolor}
\usepackage{makecell}
\usepackage{textcomp}
\usepackage{gensymb}
\usepackage{amsmath,lipsum}
\usepackage{mathtools}
\usepackage{url}
\usepackage{verbatim}
\usepackage{booktabs}
\usepackage{hhline}
\usepackage{graphicx}  % 'kotex' removed (no Korean text in body; not installed in this build env)
\usepackage[compress]{cite}
\usepackage[inkscapeformat=png]{svg}
\usepackage{wasysym} % for the diameter command 

\begin{document}

\title{\vspace*{20pt}\huge Ultra-Low-Impedance Robotic Gripper for\\ \mbox{High-Bandwidth and Transparent Physical Interaction}}
        
\author{Joon~Lee, Ari~Choi, and~Seokhwan~Jeong*

\thanks{J. Lee and A. Choi contributed equally to this work. J. Lee, A. Choi, and S. Jeong are with the Department of Mechanical Engineering, Sogang University, Seoul, South Korea. S. Jeong is the corresponding author. (e-mail: hnj1208@naver.com; cheg1541@naver.com; seokhwan@sogang.ac.kr). The mechanism presented in this work is patent pending (KR 10-2026-0005274, filed Jan. 12, 2026; PCT/KR2026/005769, filed Apr. 30, 2026).}}

% \markboth{Journal of \LaTeX\ Class Files,~Vol.~14, No.~8, August~2015}%
% {Shell \MakeLowercase{\textit{et al.}}: Bare Demo of IEEEtran.cls for IEEE Journals}

\IEEEoverridecommandlockouts % conference 모드에서 thanks를 허용하는 명령어

\maketitle

\begin{abstract}
Conventional robotic grippers often use high-ratio transmissions to generate grasping torque and external force sensors to measure physical interaction. High-ratio transmissions increase friction, reflected inertia, and mechanical impedance, while external sensors add hardware complexity. To address these trade-offs, this study proposes a novel 9-DOF, three-fingered Differential Direct-Drive (DDD) gripper that combines DD motors with a low-ratio (1:2) differential transmission. The mechanism centralizes actuator mass at the base to minimize moving-link inertia, while the differential architecture couples two motors in parallel to amplify torque during flexion. Experiments show that the prototype delivers a nominal grasping force of approximately 18 N and a fingertip force of 4.7 N, while maintaining a low motor contribution to system inertia (0.236\%) and low passive mechanical impedance, with a maximum measured value of 50.1 N/m when the motors are unpowered. The proposed hardware addresses the trade-offs among torque, physical transparency, and kinematic dexterity, providing a foundation for high-bandwidth interaction and sensorless proprioceptive force estimation.\end{abstract}


\IEEEpeerreviewmaketitle

\section{Introduction} 
\IEEEPARstart{D}{exterous} manipulation, such as grasping and in-hand manipulation, inherently involves physical interaction with unstructured environments. Managing these interactions effectively requires a system that is intrinsically responsive to contact force\cite{hogan1984impedance}. Therefore, to realize robust and adaptive manipulation, hardware with ultra-low inertia and high physical transparency is needed.

While conventional grippers often employ high-gear-ratio actuators for torque generation, the resulting mechanical nonlinearities, such as friction and backlash, degrade the system's physical transparency. Because this degraded transparency distorts contact dynamics, conventional systems often rely on external force sensors to perceive environmental interactions. Although such sensors provide direct force measurements, they add hardware complexity and can introduce sensing and filtering dynamics into the force-control loop\cite{eppinger1992three}.

To fulfill this critical requirement of high physical transparency, researchers have turned to highly backdrivable actuator paradigms. Direct-Drive (DD) motors have been employed to completely bypass the structural constraints of gear reduction \cite{bhatia2019direct}. While DD systems offer transparent power transmission, resulting in high control bandwidth and the capability for sensorless force estimation, they suffer from low torque output. Alternatively, Quasi-Direct-Drive (QDD) actuators have been proposed to secure both sufficient grasping force and proprioception\cite{gealy2019quasi}. Yet, QDD grippers still lack the physical transparency of DD systems and the kinematic degrees of freedom (DOFs) essential for dexterous manipulation.

Motivated by these trade-offs among torque, transparency, and kinematics, this study proposes a novel Differential Direct-Drive (DDD) actuation mechanism for a 9-DOF, three-fingered gripper that integrates DD motors with differential gear transmission. This architecture optimizes actuator placement to provide multiple DOFs while compensating for the torque-density limitations of pure DD systems. Furthermore, the proposed mechanism maximizes physical transparency, enabling external-force estimation based solely on motor current while retaining dexterous motion capability.

\begin{figure}
    \centering
    \includegraphics[width=1\linewidth]{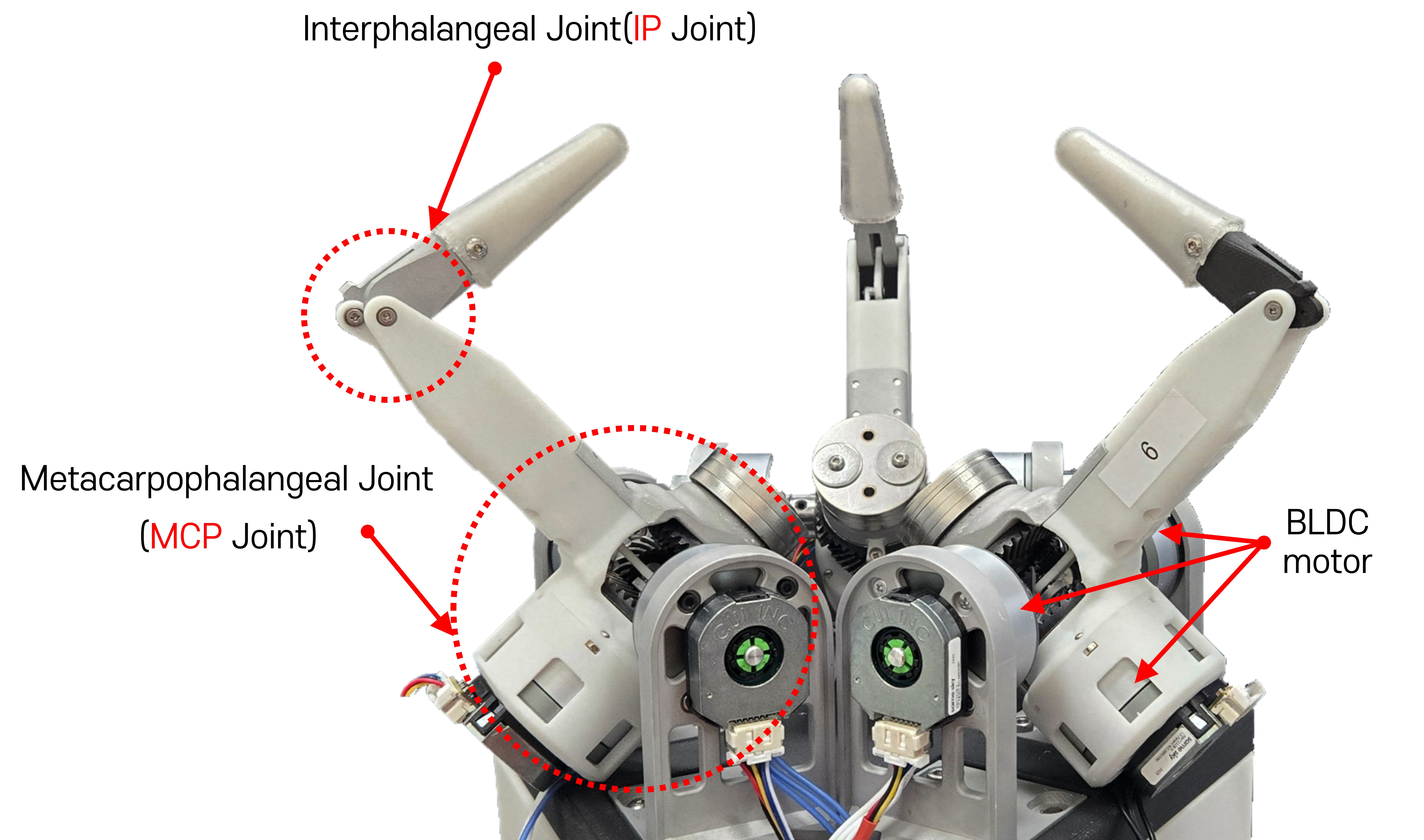}
    \caption{Overall View of the Differential Direct-Drive Gripper}
    \label{fig:DDDOverallView}
\end{figure}
%%%%%%%%%%%%%%%%%%%%%%%%%%%%%%%%

\section{Mechanism Design} \label{sec:Design Concept}
\par 
This study presents the mechanism of the Differential Direct-Drive (DDD) gripper, which utilizes a total of nine DD motors—three dedicated to each finger—coupled with a differential gear transmission. Fig. \ref{fig:DDDOverallView} illustrates the overall hardware configuration of the proposed DDD gripper. The gripper comprises three fingers, each with two mechanical joints: the metacarpophalangeal (MCP) and interphalangeal (IP) joints. Each finger provides three DOFs: MCP flexion, MCP abduction, and IP flexion.

\begin{figure}
    \centering
    \includegraphics[width=1\linewidth]{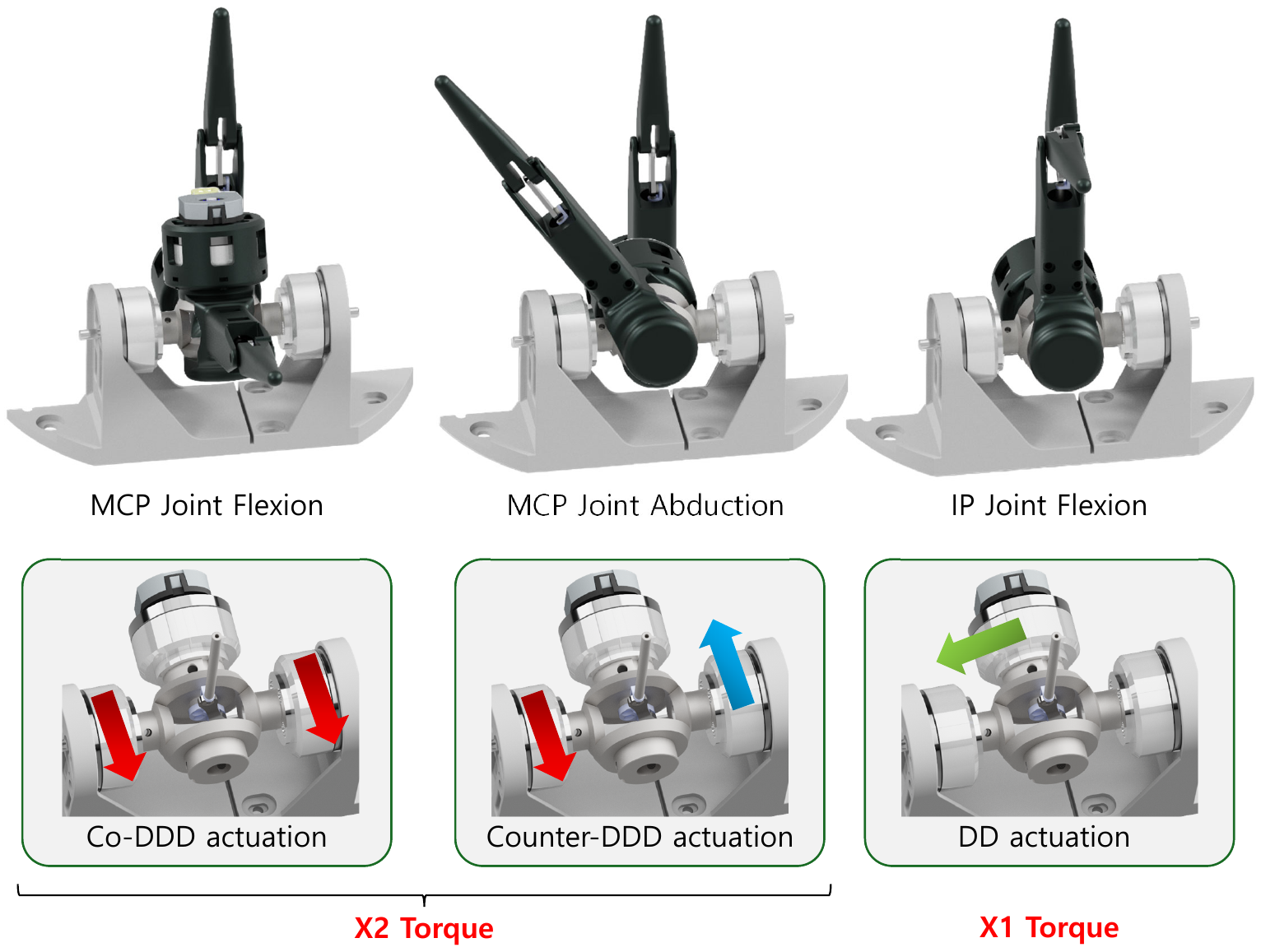}
    \caption{Operating Principle}
    \label{fig:OperatingPrinciple}
\end{figure}

As shown in Fig. \ref{fig:OperatingPrinciple}, the MCP joint is actuated by two motors via the differential mechanism. Synchronous rotation of both motors in the same direction produces flexion, whereas rotation in opposite directions results in abduction. The IP joint is driven by the rear motor utilizing a four-bar linkage mechanism. By restricting the linkage to operate within its linear region, the system avoids kinematic singularities while ensuring a linear, 1:1 angular mapping between the actuator and the joint.

The integration of the differential gear mechanism serves critical purposes in addressing the inherent limitations of pure DD actuators by optimizing both mass distribution and torque output. First, centralizing the relatively heavy motor modules at the base of the finger achieves an ultra-low inertia design, which significantly minimizes moving link inertia and enhances dynamic responsiveness. Second, it strategically allocates actuation power. As the MCP joint requires higher torque than the IP joint due to a larger effective moment arm during object grasping, the differential mechanism couples two motors in parallel during MCP flexion, providing a twofold (2x) torque amplification. To mitigate the friction typically introduced by geared transmissions, the differential employs a low gear ratio of 1:2. By leveraging this DD foundation, the system supports high-bandwidth control while preserving backdrivability and physical transparency for sensorless force estimation.

%%%%%%%%%%%%%%%%%%%%%%%%%%%%%%%%%

\section{Implementation} \label{sec:Fabrication}
\par 
The prototype of the DDD gripper utilizes BLDC motors (GL35, CubeMars). The overall structural frame is fabricated via 3D printing (Form 4, Formlabs). A single finger module weighs approximately 800 g, resulting in a total gripper weight of 2.4 kg. Since placing the DD motors near the MCP joint shifts the center of mass, integrated counterweights are employed. This mass distribution was kinematically optimized to mitigate weight imbalance, thereby reducing the gravity-compensation burden on the controllers and improving equilibrium during dynamic motion.

Performance evaluations validate the efficacy of this design. In terms of force generation, the system delivers a fingertip force of 4.7 N per module, a total nominal grasping force of approximately 18 N, and a payload of approximately 2 kg in a three-finger grasp. In terms of dynamic performance, the motors' contribution to the total system inertia is 0.236\%. With the motors unpowered, the passive mechanical impedance, quantified as $|F/x|$, reaches a maximum measured value of 50.1 N/m across the tested axes: 50.1 N/m for MCP abduction (the stiffest axis), 15.6 N/m for MCP flexion, and 9.5 N/m for PIP flexion.

\begin{figure}
    \centering
    \includegraphics[width=1\linewidth]{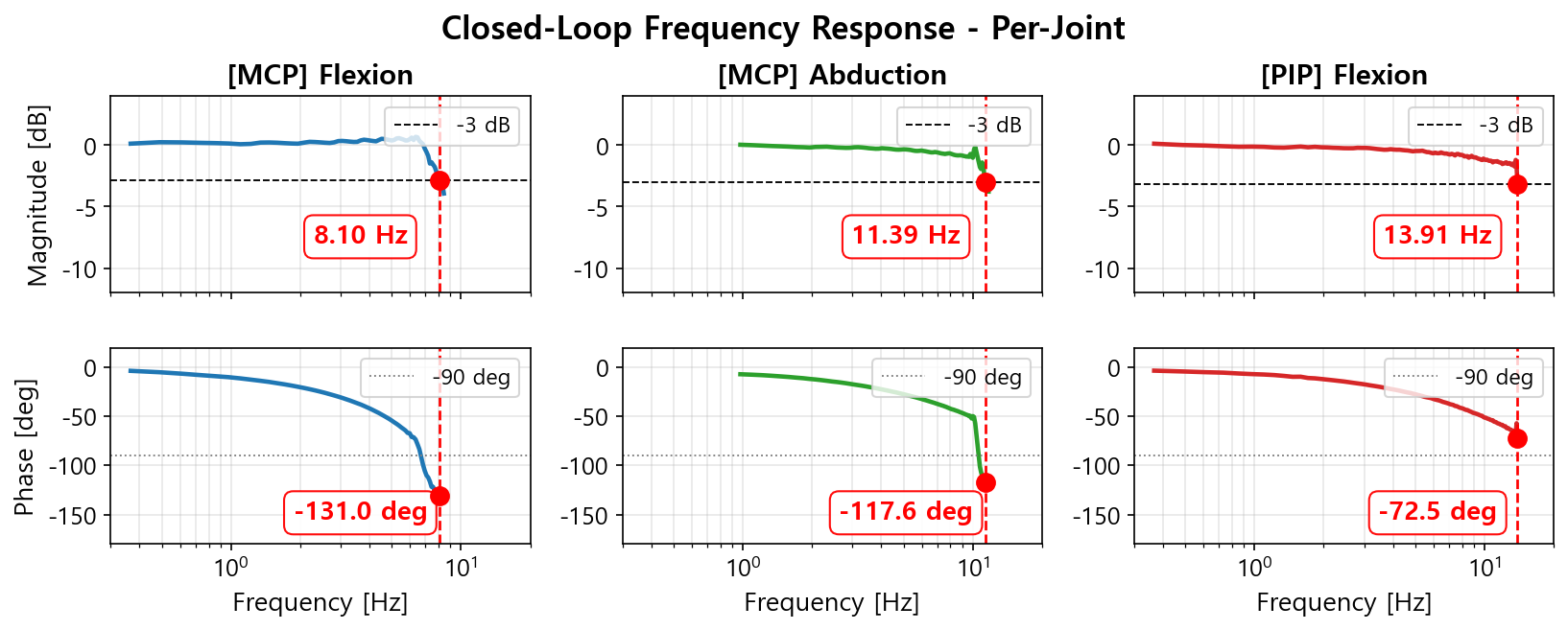}
    \caption{Joint position control bandwidth: (a) MCP joint flexion, (b) MCP joint adduction/abduction, and (c) PIP joint flexion.}
    \label{fig:joint_bandwidth}
\end{figure}

To quantitatively verify that the proposed DDD mechanism operates on a human-like time scale, the closed-loop position control bandwidths of the MCP and PIP joints were measured. Under a chirp reference input ($0.1$--$30\,\text{Hz}$), the joints were excited within their respective ranges of motion: $0^\circ$--$90^\circ$ for MCP/PIP flexion and $\pm30^\circ$ for MCP adduction/abduction. Joint angles were mapped to and from motor rotation angles via the drive transformation matrix. As shown in Fig. \ref{fig:joint_bandwidth}, the measured $-3\text{ dB}$ bandwidths are $8.1\text{ Hz}$ for MCP flexion, $11.38\text{ Hz}$ for MCP adduction/abduction, and $13.91\text{ Hz}$ for PIP flexion. Consistent with the inertia distribution of the differential structure, MCP flexion exhibits the slowest response due to the maximum accumulated equivalent rotational inertia from co-directional motor torques, while the direct-drive PIP flexion yields the fastest response.

%%%%%%%%%%%%%%%%%%%%%%%%%%%%%%%%%%%%%5

\section{Discussion and Conclusion}

\par 
In this study, we developed a novel 9-DOF Differential Direct-Drive (DDD) gripper designed to resolve the inherent trade-offs among torque output, physical transparency, and kinematic dexterity. By integrating DD motors with a low-ratio differential mechanism, the gripper successfully achieves an ultra-low inertia design while maintaining a sufficient nominal grasping force. These hardware characteristics demonstrate the system's potential for high-bandwidth control and robust physical interaction without relying on external force sensors.

Building upon the high physical transparency achieved by the proposed mechanism, future work will primarily focus on developing a robust proprioceptive force estimation framework. By leveraging the system's low inertia and high backdrivability, we plan to implement sensorless force control algorithms based on accurate dynamic modeling. While further structural optimization of the frame will be conducted as a secondary refinement, our ultimate goal is to demonstrate advanced, sensorless dexterous manipulation in dynamic real-world environments.

\bibliographystyle{IEEEtran}
\bibliography{References}

@inproceedings{bhatia2019direct,
  title={Direct drive hands: Force-motion transparency in gripper design},
  author={Bhatia, Ankit and Johnson, Aaron M and Mason, Matthew T},
  booktitle={Robotics: science and systems},
  year={2019}
}

@article{eppinger1992three,
  title={Three dynamic problems in robot force control},
  author={Eppinger, Steven D and Seering, Warren P},
  journal={IEEE Transactions on Robotics and Automation},
  volume={8},
  number={6},
  pages={751--758},
  year={1992},
  publisher={IEEE}
}

@inproceedings{gealy2019quasi,
  title={Quasi-direct drive for low-cost compliant robotic manipulation},
  author={Gealy, David V and McKinley, Stephen and Yi, Brent and Wu, Philipp and Downey, Phillip R and Balke, Greg and Zhao, Allan and Guo, Menglong and Thomasson, Rachel and Sinclair, Anthony and others},
  booktitle={2019 international conference on robotics and automation (ICRA)},
  pages={437--443},
  year={2019},
  organization={IEEE}
}

@inproceedings{hogan1984impedance,
  title={Impedance control: An approach to manipulation},
  author={Hogan, Neville},
  booktitle={1984 American control conference},
  pages={304--313},
  year={1984},
  organization={IEEE}
}

\end{document}